\documentclass[manuscript,screen]{acmart}
\AtBeginDocument{%
  \providecommand\BibTeX{{%
    \normalfont B\kern-0.5em{\scshape i\kern-0.25em b}\kern-0.8em\TeX}}}

\setcopyright{acmcopyright}
\copyrightyear{2022}
\acmYear{2022}
\acmDOI{XXXXXXX.XXXXXXX}
\acmConference[]
\acmBooktitle{} 
\acmPrice{}
\acmISBN{978-1-4503-XXXX-X/18/06}

\usepackage{makecell}

\begin{document}

\title{Robotic Maintenance of Road Infrastructures: The HERON Project}

\author{Iason Katsamenis}
\affiliation{
  \institution{Institute of Communications and Computer Systems}
  \city{Athens}
  \country{Greece}}

\author{Matthaios Bimpas}
\affiliation{
  \institution{Institute of Communications and Computer Systems}
  \city{Athens}
  \country{Greece}}

 \author{Eftychios Protopapadakis}
 \affiliation{
   \institution{Institute of Communications and Computer Systems}
   \city{Athens}
   \country{Greece}}

 \author{Charalampos Zafeiropoulos}
 \affiliation{
   \institution{Institute of Communications and Computer Systems}
   \city{Athens}
   \country{Greece}}

 \author{Dimitris Kalogeras}
 \affiliation{
   \institution{Institute of Communications and Computer Systems}
   \city{Athens}
   \country{Greece}}

 \author{Anastasios Doulamis}
 \affiliation{
   \institution{Institute of Communications and Computer Systems}
   \city{Athens}
   \country{Greece}}

 \author{Nikolaos Doulamis}
 \affiliation{
   \institution{Institute of Communications and Computer Systems}
   \city{Athens}
   \country{Greece}}

 \author{Carlos Martín-Portugués Montoliu}
 \affiliation{
   \institution{ACCIONA Construcción SA}
   \city{Madrid}
   \country{Spain}}

 \author{Yannis Handanos}
 \affiliation{
   \institution{Olympia Odos Operation SA}
   \city{Megara}
   \country{Greece}}

 \author{Franziska Schmidt}
 \affiliation{
   \institution{Université Gustave Eiffel}
   \city{Marne la Vallée}
   \country{France}}

 \author{Lionel Ott}
 \affiliation{
   \institution{ETH Zurich}
   \city{Zurich}
   \country{Switzerland}}

 \author{Miquel Cantero}
 \affiliation{
   \institution{Robotnik Automation SLL}
   \city{Valencia}
   \country{Spain}}

 \author{Rafael Lopez}
 \affiliation{
   \institution{Robotnik Automation SLL}
   \city{Valencia}
   \country{Spain}}

\renewcommand{\shortauthors}{I. Katsamenis, et al.}

\begin{abstract}
Of all public assets, road infrastructure tops the list. Roads are crucial for economic development and growth, providing access to education, health, and employment. The maintenance, repair, and upgrade of roads are therefore vital to road users' health and safety as well as to a well-functioning and prosperous modern economy. The EU-funded HERON project will develop an integrated automated system to adequately maintain road infrastructure. In turn, this will reduce accidents, lower maintenance costs, and increase road network capacity and efficiency. To coordinate maintenance works, the project will design an autonomous ground robotic vehicle that will be supported by autonomous drones. Sensors and scanners for 3D mapping will be used in addition to artificial intelligence toolkits to help coordinate road maintenance and upgrade workflows.
\end{abstract}



\maketitle

\section{Introduction}

\begin{figure}[ht]
\centering
\includegraphics[scale=0.18]{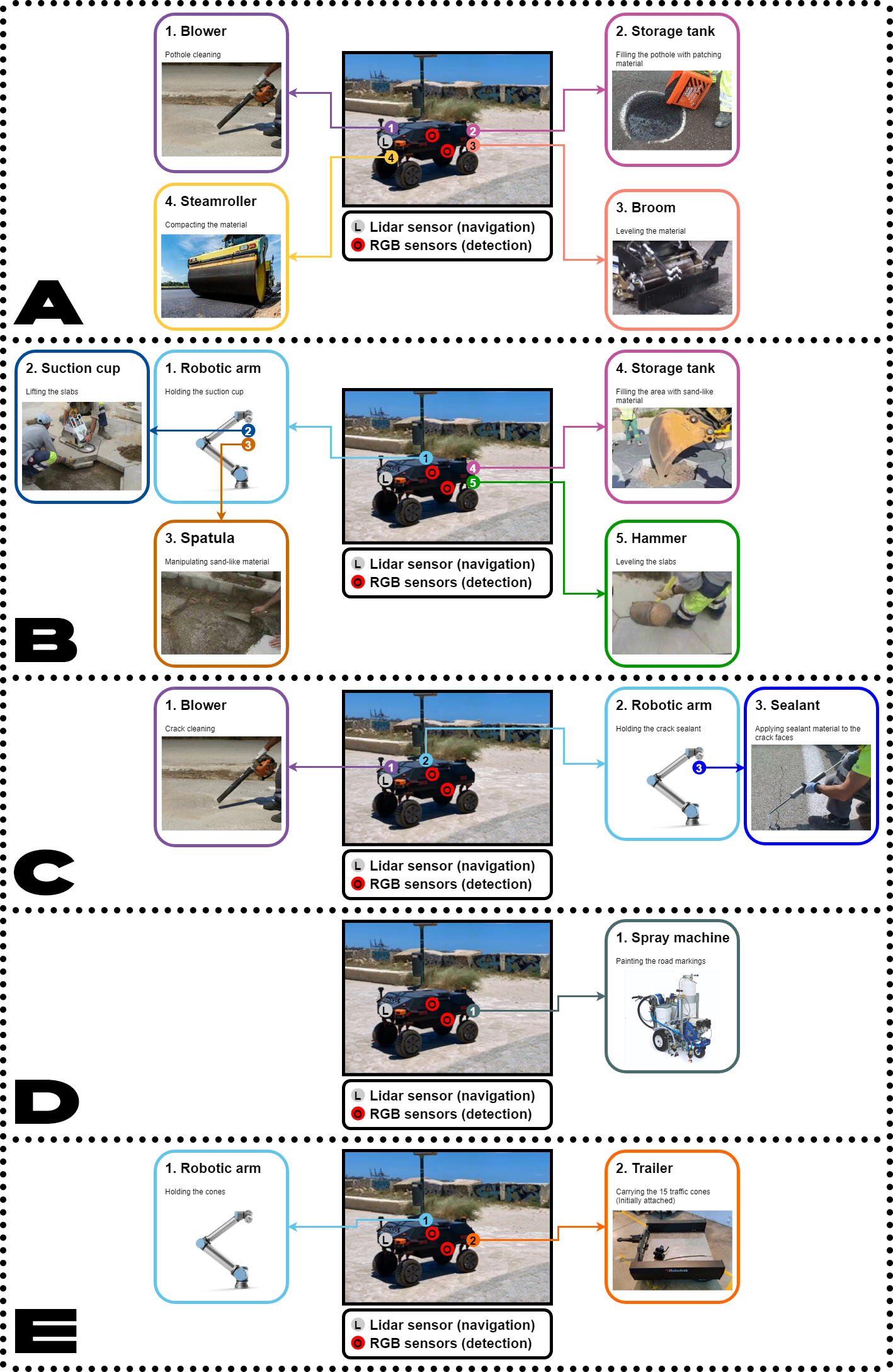}
\caption{The main maintenance procedures that will be performed by the HERON system, i.e., (A) patching potholes, (B) replacement of CUD elements, (C) sealing cracks, (D) painting blurred road markings, and (E) dispensing and removing traffic cones in an automated and controlled manner.}
\label{WP3_4_5_Diagram}
\end{figure}

Nowadays, the fast and efficient inspection, assessment, maintenance and safe operation of existing road infrastructures including highways and the overall Road Infrastructure (RI) network transport constitute a great challenge for many operators and engineers \cite{katsamenis2020pixel}. HERON aims to develop an integrated automated system to perform maintenance and upgrading roadworks tasks, such as sealing cracks (see Fig. ~\ref{WP3_4_5_Diagram}C), patching potholes (see Fig. ~\ref{WP3_4_5_Diagram}A), asphalt rejuvenation, autonomous replacement of CUD (removable urban pavement) elements (see Fig. ~\ref{WP3_4_5_Diagram}B) and painting road markings (see Fig. ~\ref{WP3_4_5_Diagram}D), but also supporting the pre- and post-intervention phase including visual inspections and dispensing and removing traffic cones (see Fig. ~\ref{WP3_4_5_Diagram}E) in an automated and controlled manner \cite{loupos2018autonomous}.

\begin{figure*}[ht]
\centering
\includegraphics[width=\linewidth]{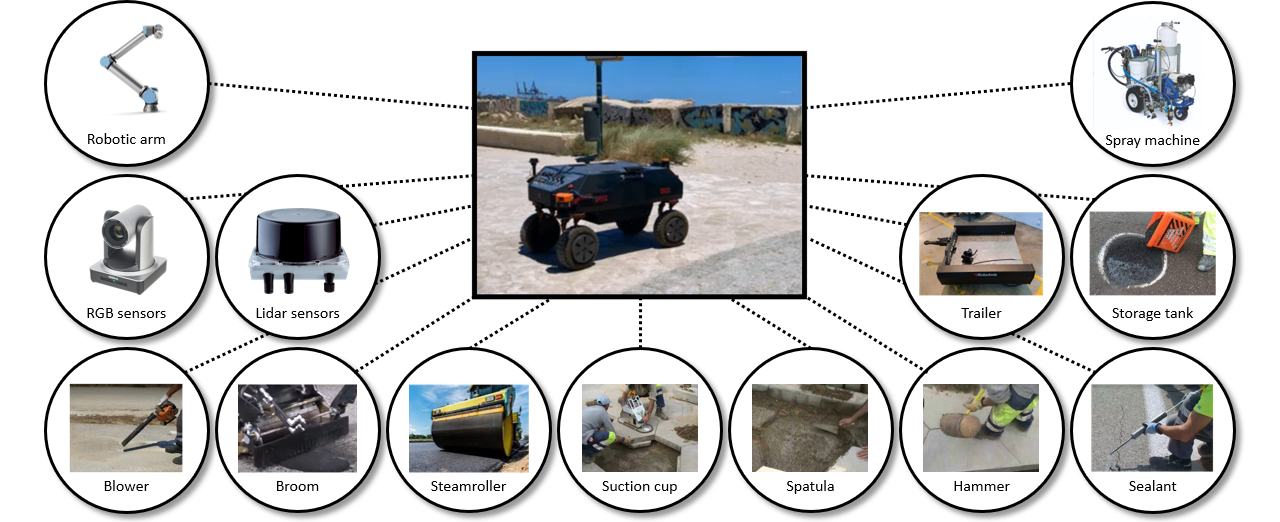}
\caption{From traditional tools to robotic sensors and actuators.}
\label{sensors_actuators}
\end{figure*}

The HERON system consists of: i) autonomous ground robotic vehicle that will be supported by autonomous drones to coordinate maintenance works and the pre-/post-intervention phase \cite{galar2020robots}; ii) various robotic equipment (see Fig. ~\ref{sensors_actuators}), including sensors and actuators (e.g., tools for cut and fill, surface material placement and compaction, modular components installation, laser scanners for 3D mapping) placed on the main vehicle \cite{bock2016construction}; iii) sensing interface installed both to the robotic platform and to the RI to allow improved monitoring (situational awareness) of the structural, functional and RI’s and markings’ conditions \cite{protopapadakis2020multi}; iv) the control software that interconnects the sensing interface with the actuating robotic equipment \cite{protopapadakis2015image}; v) Augmented Reality (AR) visualization tools that enable the robotic system to see in detail surface defects and markings under survey \cite{carneiro2018bim}; vi) Artificial Intelligence/AI-based toolkits that will act as the middleware of a twofold role for: a) optimally coordinating the road maintenance/upgrading workflows and b) intelligent processing of distributed data coming from the vehicle and the infrastructure sensors for safe operations and not disruption of other routine operations or traffic flows \cite{varona2020deep, katsamenis2020man, s20216205, lv2014traffic}; vii) integrate all data in an enhanced visualisation user interface supporting decisions \cite{zudilova2009overview, katsamenis2022fall}; and viii) communication modules to allow for Vehicle-to-Infrastructure/Everything data exchange for predictive maintenance and increase users' safety \cite{malik2019mapping}. Consequently, HERON will allow for a modular design of the system operation, maximizing its capabilities and adaptability for various transport infrastructures, while reducing fatal accidents, maintenance costs, and traffic disruptions, thus increasing the network capacity and efficiency.

\section{Related work}

HERON is one of the EU-funded, under H2020, projects for the Road infrastructure maintenance program. HERON's part includes the execution of a road infrastructure plan created by related projects. To this end, HERON aims at innovative engineering of links and connections to allow a smooth transfer from one mode to another in case of extreme disruption in one transport mode.


Many other projects, like HERON, have been funded by the EU to an autonomous ground robotic vehicle, supported by autonomous drones, that will provide better monitoring, assessment, and maintenance of road infrastructures. The EU-funded InfraROB \cite{InfraRob} project will focus on automating, robotising, and modularising road construction and maintenance work. More specifically, it will develop, among others, autonomous robotized systems and machinery to carry out line marking, repaving, and the repair of cracks and potholes. It will also develop collaborative robotized safety systems for construction workers and road users. The project further aims to integrate pavement management system and traffic management system solutions for a holistic, unified management of road infrastructure and live traffic \cite{gonzalez2022detection}.  

The EU-funded OMICRON \cite{Omicron} project will develop an intelligent asset management platform (IAMP) with a vast portfolio of area-specific innovative technologies to increase the construction, maintenance, renewal, and upgrade of the EU road network. The project will address the entire road network system, focusing on digital inspection technologies implementation, road digital twin development, construction of a decision support tool, intelligent construction development, and intervention solution for infrastructures. The IAMP will be interconnected by a building information modelling (BIM) oriented digital twin to the decision support tool to enable industrialisation and automation of several road management functions, demonstrated in Italy and Spain. 

\begin{figure}[ht]
\centering
\includegraphics[width=\linewidth]{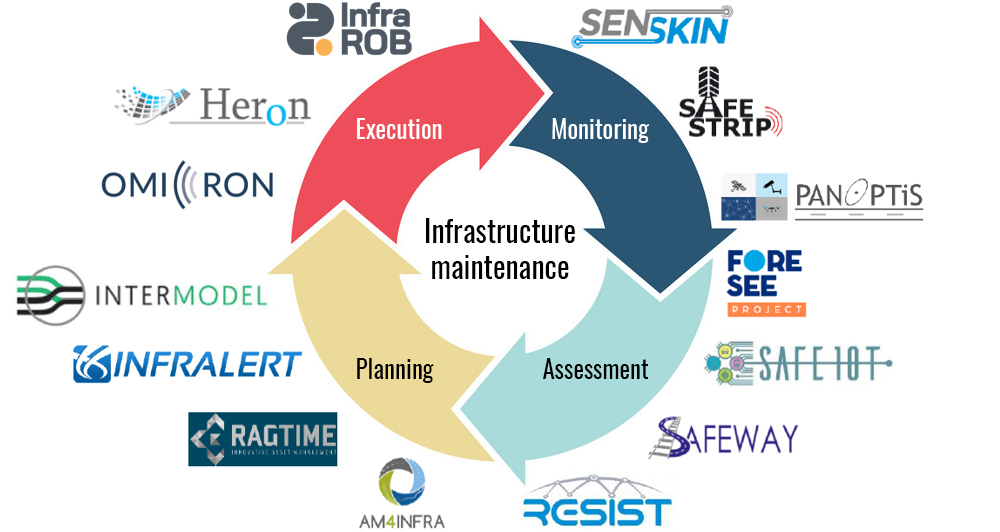}
\caption{HERON is one of the three Research and Innovation Actions that close the maintenance cycle.}
\label{maintenance_cycle}
\end{figure}

Lastly, the PANOPTIS \cite{sevilla2018improving} project aims at increasing the resilience of the road infrastructures and ensuring reliable network availability under unfavourable conditions, such as extreme weather, landslides, and earthquakes. Its main target is to combine downscaled climate change scenarios (applied to road infrastructures) with simulation tools (structural/geotechnical) and actual data (from existing and novel sensors), so as to provide the operators with an integrated tool able to support more effective management of their infrastructures at planning, maintenance, and operation level. All the aforementioned projects, along with HERON, are funded by the European Union's Horizon 2020 Research and Innovation Programme. It is noted that HERON is one of the three Research and Innovation Actions funded under the topic of ameliorating environmental impacts and full automated infrastructure upgrade and maintenance of the Horizon 2020 program SOCIETAL CHALLENGES - Smart, Green and Integrated Transport. (see Fig. ~\ref{maintenance_cycle}).

\section{The HERON project}

\begin{figure*}[ht]
\centering
\includegraphics[width=\linewidth]{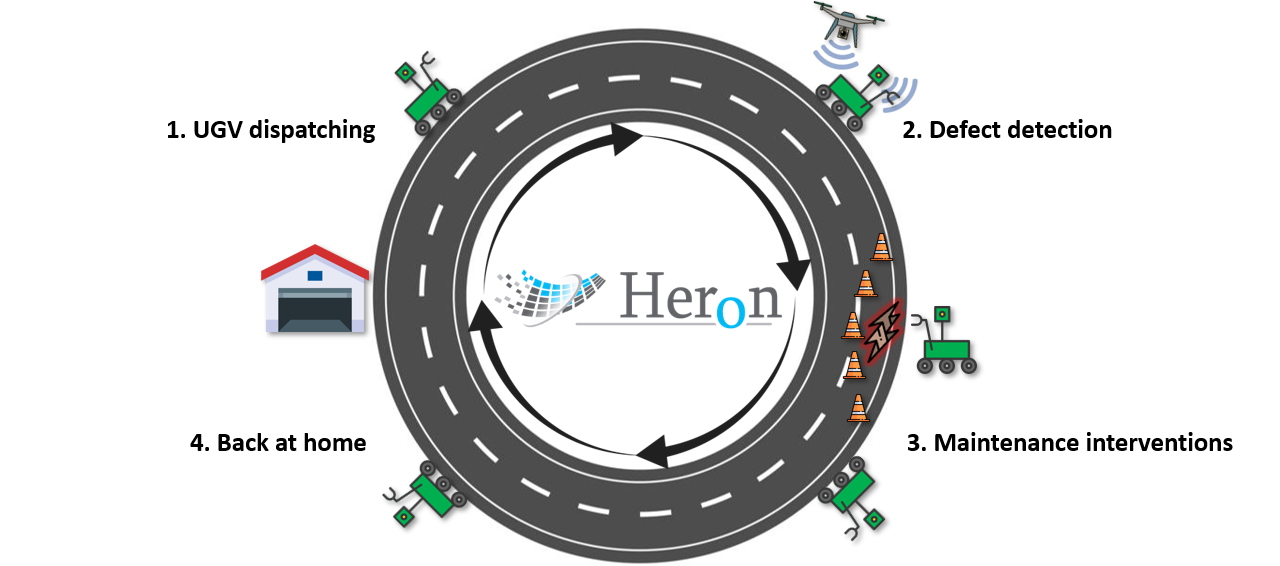}
\caption{HERON’s concept.}
\label{concept}
\end{figure*}

The (semi-)automated HERON system relies on improved intelligent control of a multi-degree-of-freedom (MDOF) robotized vehicle, improved computer vision (CV), and Artificial Intelligence (AI)/Machine Learning (ML) techniques \cite{katsamenis2022simultaneous} combined with proper sensors (see Fig. ~\ref{sensors_actuators}), decision-making algorithms and AR components to perform corrective and preventive maintenance and upgrading of roadworks is considered an advanced solution, which pushes routine roadwork activities quite beyond the state-of-the-art. At the same time, by using advanced data coming from various sources (including V2I and aerial drone surveillance) and well-established methods (from existing know-how from research and industrial projects), the automated system will be able to provide some non-routine (emergency) maintenance operations when required. Towards that direction, HERON targets the development and prototype validation of an innovative, automated intelligent robotic platform for performing the above tasks safely, promptly, reliably, and modularly (see Fig. ~\ref{concept}).

Towards this direction, HERON targets the following seven Scientific and Technical Objectives. First, HERON aims to develop an integrated robotic platform with increased navigation and positioning capabilities for maintenance and upgrading tasks and concurrent assessment of the RIs. HERON improves Real-Time (RT) perception and cognition of the robotic vehicles, through Simultaneous Localisation and Mapping (SLAM) and novel Machine Learning techniques \cite{bakalos2021man, frasson2021unsupervised}. SLAM will provide the concurrent mapping generation of the robotic vehicles’ space (complementary to GPS). The algorithms will be used for the navigation of the robotized vehicles allowing manoeuvres and avoiding collisions. The improved cognition will be based on recent advances in Computer Vision and ML for detecting objects, signalling, road markings, and Points of Interest (PoIs), interpreting events and actions, which occur in the area of intervention. Existing robotic vehicles will be used and enhanced with all necessary sensors and cognition devices to perform the maintenance and assessment of the RI \cite{galar2020robots, skibniewski1990automation}.

Secondly, this project will provide an optimized control framework for developing and refining robotic manipulation skills required to perform assessment and maintenance, and upgrading interventions of RI \cite{kalashnikov2018scalable, doliotis20163d}. This requires: a) specialized robotic equipment attached to the vehicle to take care of certain RI maintenance and upgrading works such as modular components installation, cut and filling, surface material placement and compaction, painting, spraying, and b) a control mechanism to transform the high-level processes into low-level robotic manipulation skills \cite{erdem2011combining}. These tools are in close relation with the robotic equipment programming algorithms and the sensing interface. Improved learning-based methods to acquire specialized low-level manipulation and interaction abilities will be developed (e.g., grasping, pulling, pushing, cutting, filling) through a combination of sim-to-real and user-provided demonstrations. These low-level abilities will be further refined into higher-order skills (e.g., pouring, levelling, etc.) by leveraging a symbolic representation framework. Using such representations provides access to powerful planning tools that can be used to generate high-level behaviours such as sealing cracks, patching potholes, painting markings, etc \cite{schmid2020efficient}.

\begin{figure}[ht]
\centering
\includegraphics[width=\linewidth]{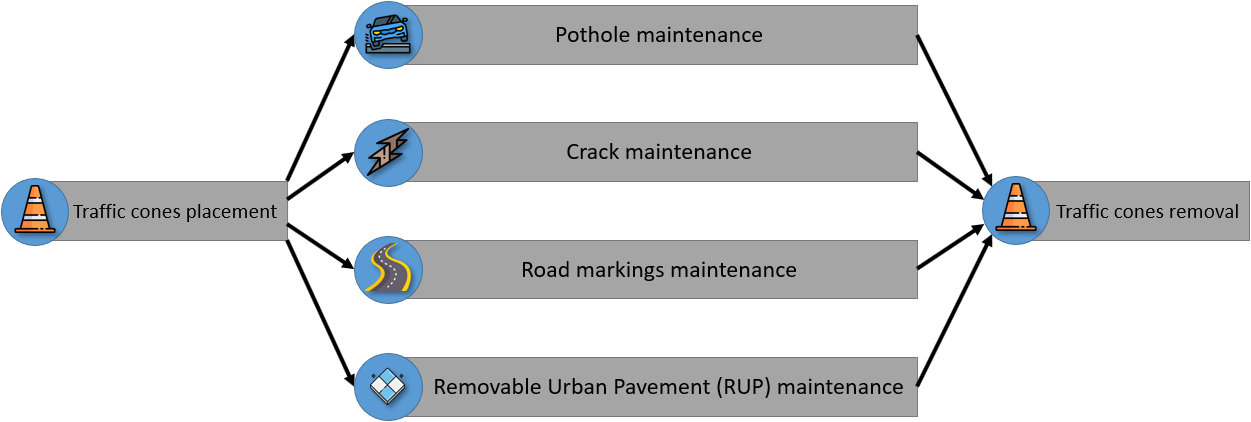}
\caption{Identified use cases of the HERON project.}
\label{identified_use_cases}
\end{figure}

Thirdly, HERON will integrate improved sensing and communication capabilities to the robotic platform in order to extract the required measurements in the identified areas of concern (e.g., poor or lack of markings, surface defects and potholes, dispensing and removing traffic cones, etc.) within an acceptable degree accuracy (see Fig. ~\ref{identified_use_cases}). In order to achieve a precise, automated maintenance and upgrading task, an accurate inspection of the PoI is needed, as well as wider knowledge of the surrounding area. This is done through the sensing interface installed on the robotic vehicle, the drones, and on the RI. Surface 3D mapping and modelling are some of the salient parts of HERON sensors aiming at extracting precise details on the 3D geometry of a road defect and thus stimulate the proper control mechanisms for its repair or the activation of pre-fabricated solutions (such as the Demountable Urban Roadway -CUD- elements) to modularize the planning and accelerate the maintenance and upgrading tasks \cite{zhao20173d, oude20093d, zai20173, elberink20063d}. CV tools will be applied towards a traffic management for a safe flow and working conditions of the maintenance personnel (access to PANOPTIS outcomes) \cite{buch2011review}. Regarding the communication tools, HERON will support any module (e.g., 4G/5G, WiMAX, BLE4, etc.) that will be integrated into the robotic vehicle, allowing the robotized vehicle(s) to have different configurations for communicating to the each other, the field maintenance team and the control center \cite{ma2017gan}.

The fourth scientific objective includes the implementation of an AI toolkit enriched with image analysis modules to optimally coordinate the whole maintenance process and simultaneously process in a smart way the data from the sensing interface to take accurate and prompt decisions which can guarantee an unhindered execution of the traffic flow and a safe execution of routine maintenance works by the personnel. The AI tools will be developed on novel deep machine learning algorithms and receive as input information from all data sources, installed either on the robotic platform or in the RI or being inputted from external sources (proper V2I data exchange for predictive maintenance, Copernicus, etc.), fuse this information and extract high-level conceptual decisions on the maintenance process. In addition, image analysis and CV algorithms will be applied to allow a continuous traffic flow during the maintenance works and guarantee a safe execution of routine works on the RI by the expert-personnel. Convolutional Neural Networks (CNNs), Long Short-Term Memory Networks (LSTMs), and Generative Adversarial Networks (GANs) will be exploited as Deep Learning (DL) tools, which will be aligned with vision-based object detection algorithms, defect detection schemes, and traffic flow analysis modules \cite{voulodimos2018deep, ye2021convolutional, mansouri2019reducing, pei2021virtual}. Finally, HERON supports a distributed AI architecture including the widely publicized paradigm of Federated Learning, which is the epitome of collaboration between edge nodes (Unmanned Ground Vehicles-UGVs) and cloud for meeting privacy and application performance requirements. HERON will also focus on optimizing a cloud/edge solution for a big class of robotic applications in RIs.

The next scientific objective refers to the design and development of proper communication architecture to support seamless and ubiquitous services among the various actors of the maintenance and upgrading operations (sensors, UGVs, and drones). Robust communication will be developed for keeping the network organized even in case of a link/node failure; the HERON network will be self-healing with reconfiguration around a broken path or alternative connection, ensuring continuous connectivity. Our proposed architecture aims to operate in a seamless and cooperative manner with existing human maintenance and technical crews of the roads, as well as integrate HERON into the existing maintenance/upgrading mechanisms. It will act as the technological backbone to provide improved Data Fusion (DF), seamless interconnection, and interoperability between the system layers minimizing data ambiguity \cite{du2020real, mohd2022multi}.

The sixth scientific objective is the implementation of an integrated Decision Support System (DSS) and an advanced Incident Management System (IMS) with interactive AR/VR visualization tools \cite{sharda1988decision, allred2004coordinating, riegler2021systematic}. HERON will integrate all the information to be provided by the various tools, the sensor data, etc. as different layers in a unified enhanced visualization User Interface (UI) generating the Common Operational Picture (COP), as complementary support to the prevailing Transport Management Systems (TMS) and Closed-Circuit Television (CCTV). The IMS will provide information on facilities, equipment, personnel, procedures, and communications. Assisted by the DSS tool, IMS will manage seamlessly routine incidents as well as emergency situations in the RIs. The decisions and response procedures per reported incident will be turned into actions and resource proposals by building-related incident process workflows to support the decision-making of RI operators. The visualization will be enriched with AR components to allow for the user to have in-situ supervision of the automated maintenance and upgrading operation process.

In the last scientific and technical objective, HERON aims to implement on-site integration, scaled-demonstration of the services and the technological components, and validation of the HERON platform in three case studies (Greece, France, and Spain). The cost, features, benefits, and reaction time of the proposed system and incurred procedures will be compared to conventional (pre-project) procedures and means of motorway operation.

\subsection{Action plan}

\begin{figure}[ht]
\centering
\includegraphics[scale=0.65]{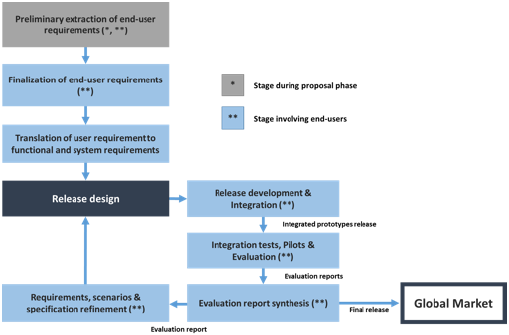}
\caption{HERON evolutionary process; from idea into market realization.}
\label{HMR}
\end{figure}

Targeting high user acceptability and a clear demonstration of its concepts at the end, HERON is clearly a data-driven research project, with strong industry involvement and with almost 6-months piloting activities scheduled, having as an ultimate goal the achievement of the highest possible TRL concerning the technology robotics and control systems, AI components, sharing features and the overall system developed in its context. This orientation is even more strengthened through the adoption of an agile development schema that aims to produce integrated prototypes early in the project. The realization of HERON will adopt a start-up mentality, targeting to successfully identify the appropriate competences that HERON needs to possess in order to achieve high technology acceptance and forming a team that has the ability to perform efficiently and deliver high-quality results to drive innovation/excellence a step further. HERON will follow an evolutionary process (common in novel product development activities) by iterating a series of activities (see Fig. ~\ref{HMR}). The proposed iterative approach will be realized in three development cycles, allowing new data from the validation studies to be incorporated into the product development process, revising it whenever necessary. Throughout the entire process, other similar initiatives will be monitored in order to identify new opportunities, threats, and potential for collaborations.  

\subsection{Design and implementation}

HERON will follow a process by iterating a series of activities, performing initial modules, system assessments, and validation campaigns well before the actual demonstrations on sites. This approach will be realized in 2 cycles, allowing new data –after evaluation- to be incorporated into the outputs development process, revising it when necessary. The 1st cycle closes with the delivery of the initial integrated HERON system. The 2nd cycle ends through the final integrated system based on the technical and operational assessments of the 1st round of demonstration. Next, HERON will adopt a fast-failure approach in terms of getting through the steps in the concept maturing and system evolution process. The project team’s peripheral vision will be used in order to keep a live roster of opportunities, threats and challenges in the area of interest, allowing an effective alignment to the current conditions throughout the HERON’s lifecycle. This will provide a means of a rigorous assessment of options and the selection of the most suitable one based on balanced trade-offs that will not hamper the overall project progress.

The approach will allow the HERON consortium to respond to external or internal opportunities during the project’s lifetime and will add to the project’s agility to accommodate really innovative solutions that will match emerging trends and needs at the actual time of implementation. Thus, the real potential of the final outputs ensures the uptake in the mid-term horizon after the end of the project. In support of its agile/iterative approach, HERON is organized to perform smaller scale intermediate validations (in collaboration with UGE and the associated Transpolis test sites), in which end-users will be introduced to the developed and evolved solutions, so as to incorporate their feedback in a timely and resource-efficient manner, aligning their needs with project outcomes to the highest possible extent. This will also be a training opportunity for the end-users (both road operators and transport authorities) so that they can understand the HERON’s value proposition in improving the safety and the resilience of RIs through the integration of the proposed hazard-awareness solutions. HERON aims to reach TRL7 as a transition from experimental to real-life conditions will be performed. This TRL is the ultimate milestone that is expected to be reached with the final prototype. The process followed will generate a gradually increasing TRL for the whole system alone and for each component individually. Fig. ~\ref{HMR} presents the technology maturity process of the HERON system and the way it is linked to each iteration and the production of prototypes.

Finally, HERON aims to use a number of methodological tools aligned with a gendered approach to innovation. First, it will ensure that an adequate number of women are participating in the research both as researchers and research participants. This is particularly important in the system development, piloting, and evaluation, where each of these tasks will need to ensure adequate numbers of women are represented. Moreover, HERON will actively support the H2020 objective to promote gender equity by ensuring that females have the same chance as male team members to participate in the project team and project events.

\subsection{Project's Impact}

The development of the HERON project will have an impact on three fundamental areas: Technical, Economic, and Occupational Safety and Health (OSH) administration. At a technical level, HERON will showcase that it is possible to make a fully functional and efficient (semi-) automated system composed of commercial machines that work in a fully integrated manner, using the latest technologies in localization, navigation, AI/ML, comprehensive planning, decision-making, and automated manipulators to carry out complex maintenance and upgrading roadworks. HERON will make use of materials that have precise and short application times (e.g., hot mixes, paints, bitumen. etc.). Economically, once the HERON technology will be technically proven and widely accepted by both road authorities and the public, then it will reach a point of commercialization. HERON is expected to have an impact in reducing the personnel needed to perform on-site roadworks. These people could be assigned to perform other tasks in the same company, which would reduce overall execution times or carry out a greater number of activities. Specifically, focusing on actions on the roadworks, a summary of the estimated reduction in resources per task is shown in the Table ~\ref{resources_reduction}. This local reduction of workers (although the total number will remain the same in the company) will make it possible to amortize the technological investment of the systems included in machines and their integration, generating economic savings from year “n” of amortization and holding jobs for people. At the OSH level, HERON allows fewer people to work on a road while there is still traffic on the rest of the lanes, which reduces the chances of being run over in the event of an accident. On the other hand, the number of people who are in contact with hot substances that emit toxic fumes (e.g., bitumen and paints) will be reduced to a minimum level. Based on the above analysis, the expected socio-economic and environmental impacts are described in the sections below.

\begin{table*}[ht]
\centering
 \begin{tabular}{||c | c | c | c||} 
 \hline
  \thead{Task} & \thead{Workers needed \\ conventionally} &  \thead{HERON estimated \\ number of workers} &  \thead{HERON Impact} \\
 \hline\hline
 Sealing cracks and patching potholes & 3 or 4 in-situ & \textbf{1 supervisor of the} & 33\% to 50\% \\ 
 Painting markings & 2 or 3 in-situ & \textbf{UGV assisted by} & Up to 50\% \\
 Visual inspections & 2 or 3 in-situ & \textbf{1 drone pilot} & Up to 50\% \\
 Cones system & 2 or 3 in-situ & \textbf{(if needed)} & Up to 50\% \\
 \hline
 \end{tabular}
\caption{Summary of the expected reduction of resources due to the usage of the HERON system.}
\label{resources_reduction}
\end{table*}

\subsubsection{HERON Socio-Economic Impact}

HERON Social-Economic (SE) impact includes safer and more resilient roads and highways (RI/TI in general), and consequently better conditions for the users. The provision of automated systems able to perform an inspection of roads/highways with increased accuracy, detail, efficiency, objectivity, repeatability, and data management capabilities will aid maintenance planning of the inspection personnel with the aid of adapted strategies based on the current conditions and highway needs. This is expected to highly contribute to a safer transport network for the users, particularly the wide public, commuters.

The second SE impact refers to the reduction of the need for highly trained personnel. The suggested system requires less personnel assigned to the inspection/maintenance task itself reducing thus the operational costs of the roads’/highways’ owners/operators as well as the inspection/maintenance companies. Furthermore, the solution is expected to reduce hourly labor in the transport environment which is really risky and very costly when having also in mind the transportation of relevant personnel to/from the PoI. Next impact refers to a safer working environment for maintenance, upgrading, inspection, and first-response teams, which are most of the time required to execute their tasks in the harsh environments of roads/highways always experiencing and facing inconvenient and not safe conditions such as passing vehicles, noise, dust and extreme temperatures (inspection/maintenance teams) and even worse conditions (smoke, chemicals, risk of collapse, etc.) when referring to first response units. This is expected to greatly improve by the usage of “expert/smart” systems, such as HERON, from the RI industry (as well as the wider transport infrastructure industry).

The next SE impact is the maintenance and upgrading of roadwork costs reduction through regular, structured, and planned operations. This is expected to be achieved through the introduction of more often, but still less expensive automated maintenance/upgrading processes. Currently, there is precise planning of regular inspections of highways; however, sometimes these are limited due to operational costs of the inspections, often to the limit that can be afforded. The proposed system will allow more targeted inspections, as it will contribute to improved predictive maintenance, whereas the whole RI, e.g., tunnel lining, pavement, will be inspected and not just some spots under a sampling procedure. This will lead to an improved maintenance schedule, targeting the forecast of future damages and the on-time intervention (predictive maintenance). This can greatly reduce infrastructure maintenance costs.

One additional SE impact is the reduction of road/highway downtime for maintenance and upgrading. The maintenance and upgrading works of roads and in particular of large highways usually requires partial or total shutdown of a sector and the traffic is usually diverted to local roads around the previous toll stops. This is not an easy task for the highway operators and requires the availability of actual road diversions for the drivers, which is not always the case. The development of an industrial-grade robotic system and its exhaustive testing into actual working/environment conditions will ensure the operation of the system without stopping the highway traffic. Another impact is also the reduction of vehicle operating costs, through structured inspection of the pavement which will, in turn, result in decreased fuel consumption, tire repair costs, vehicle maintenance, and depreciation.

The last impact is regarding the travel time savings. According to the EC Joint Research Centre (JRC), citizens from countries like Italy, Luxemburg, Greece, and Belgium spent more than 30 hours annually in road congestion. Timely maintenance of the RI may reduce journey times and is the most widely recognized economic benefit of road maintenance.

\subsubsection{HERON Environmental Impact}

HERON Environmental Impact includes the contribution to a greener environment. In particular, it is expected that through more regular and targeted road and highways structures’ maintenance process, there will be a reduction of actions/activities, which usually require significant raw materials consumption, machinery, and higher works time (which implies major traffic disruptions and associated emissions) and an increase of preventive measures and maintenance that are quick and less raw materials and machinery demanding.

Another environmental impact refers to decreased emissions. The transport sector has been identified as one of the most CO2-producing sectors. EC's goal is a total saving of 55\% of CO2 by 2030 and not “only” the 40\% target of the Paris agreement. Indeed, there is a clear aim to become the first climate-neutral continent. European roads support the transport of 81\% of passengers and 73\% of inland freight. Pavement surface influences the fuel or electricity consumption of vehicles through the rolling resistance between the road and tires riding over it. Evenness, rutting, potholes, and deteriorated joints. Scientific studies have shown that proper maintenance to replace pavement surfaces that show “bad” or under-performing surface conditions with smooth road surfaces with “good” properties would lead to fuel use reductions and lower CO2 emissions by up to 5\%. This means that an upgrade of just one-third of the road network of Europe by 2030 could lead to annual savings of 14 million tonnes of CO2, or the equivalent of removing the emissions associated with 3 million cars. HERON is focused on road maintenance and upgrades, so is directly targeting the improvement of pavements surface and as a result CO2 emissions reduction.

Emissions in the highway networks are expected to reduce through their appropriate operation following a proper inspection. This is also coupled with the expected reduced use of conventional manned inspection and maintenance techniques following the adoption of the HERON outputs. The expected reduction of traffic disruption will imply additional savings in the emissions released during traffic jams. The reduction of vehicle needs thank you to the use of drones for visual inspections, general patrolling, inventory activities among others due to HERON, will also have a positive impact on emissions. Lastly, HERON aims to achieve an increased lifetime of RI.

The HERON system and capabilities suggested will provide quicker and targeted maintenance and inspections at identified hot spots of RIs increasing the estimation of life, having in mind the impact of their hostile environment, enduring heavy stress, changing weather conditions, and extreme stresses. This is expected to enhance the structural integrity of the RI systems, hence imposing further environmental benefits associated with maintenance and preservation (which imply raw materials and energy consumption).

\section{Conclusion}

To conclude, HERON proposes an integrated automated system and an autonomous ground robotic vehicle that will be used for the maintenance of road infrastructures. Moreover, HERON will provide a continuous vehicle for infrastructure data exchange to increase user safety. This new integrated system will maximize its capabilities and adaptability for various transport infrastructures while reducing fatal accidents, maintenance costs, and traffic disruptions. Future work includes the realization of the integration of the individual sensors, actuators, and tools as well as the on-site demonstration of the HERON system.

\begin{acks}
This work has received funding from the European Union’s Horizon 2020 Research and Innovation Programme under grant agreement No 955356 (Improved Robotic Platform to perform Maintenance and Upgrading Roadworks: The HERON Approach).
\end{acks}

\pagebreak

\bibliographystyle{unsrt}
\bibliography{Heron}

\end{document}